\documentclass[letterpaper, 10 pt, conference]{ieeeconf}  

\usepackage{subfloat}
\usepackage{amsmath,amsfonts}
\usepackage{array}
\usepackage[caption=false]{subfig}
\usepackage{textcomp}
\usepackage{stfloats}
\usepackage{url}
\usepackage{verbatim}
\usepackage{graphicx}
\usepackage{cite}
\usepackage{amsmath,amssymb}
\usepackage{makecell}
\usepackage{bbding}
\usepackage{multirow} 
\usepackage{multicol} 
\usepackage{arydshln}
\usepackage{booktabs}

\usepackage{float}
\usepackage{svg}  
\usepackage{balance} 

\usepackage{color}
\usepackage[noend]{algpseudocode} 
\usepackage{CJK}

\makeatletter
\let\NAT@parse\undefined
\makeatother
\usepackage{hyperref}

\usepackage[linesnumbered,ruled,vlined]{algorithm2e}

\makeatletter
\newcommand{\removelatexerror}{\let\@latex@error\@gobble}
\makeatother

\UseRawInputEncoding
\IEEEoverridecommandlockouts                              

\UseRawInputEncoding

\title{\LARGE \bf
PKE-RRT: Efficient Multi-Goal Path Finding Algorithm Driven by Multi-Task Learning Model
}

\author{Yuan Huang$^{1}$
\thanks{*This work was supported by JST SPRING, Grant Number JPMJSP2128.}
\thanks{$^{1}$Yuan Huang is with the Graduate School of Information, Production and Systems, Waseda University, Japan
{\tt\small gakki@toki.waseda.jp}}%
}

\begin{document}

\maketitle
\thispagestyle{empty}
\pagestyle{empty}

\begin{abstract}

Multi-Goal Path Finding (MGPF) aims to find a closed and collision-free path to visit a sequence of goals orderly. As a physical travelling salesman problem, an undirected complete connected graph with accurate weights is crucial for determining the visiting order. Lack of prior knowledge of the local path between vertices poses challenges in meeting the optimality and efficiency requirements of algorithms. In this study, a multi-task learning model, designated Prior Knowledge Extraction (PKE), is designed to estimate the local path length between pairwise vertices as the weights of the graph. Simultaneously, a promising region and a guideline are predicted as heuristics for the path-finding process. Utilizing the outputs of the PKE model, a variant of Rapidly-exploring Random Tree (RRT) is proposed known as PKE-RRT. It effectively tackles the MGPF problem by a local planner incorporating a prioritized visiting order, which is obtained from the complete graph. Furthermore, the predicted region and guideline facilitate efficient exploration of the tree structure, enabling the algorithm to quickly provide a sub-optimal solution. Extensive numerical experiments demonstrate the outstanding performance of the PKE-RRT for MGPF problem with a different number of goals, in terms of calculation time, path cost, sample number, and success rate.
\end{abstract}

\section{INTRODUCTION}
\label{sec1}
\begin{figure}[t]
	\centering
	\includegraphics[width=3.2 in]{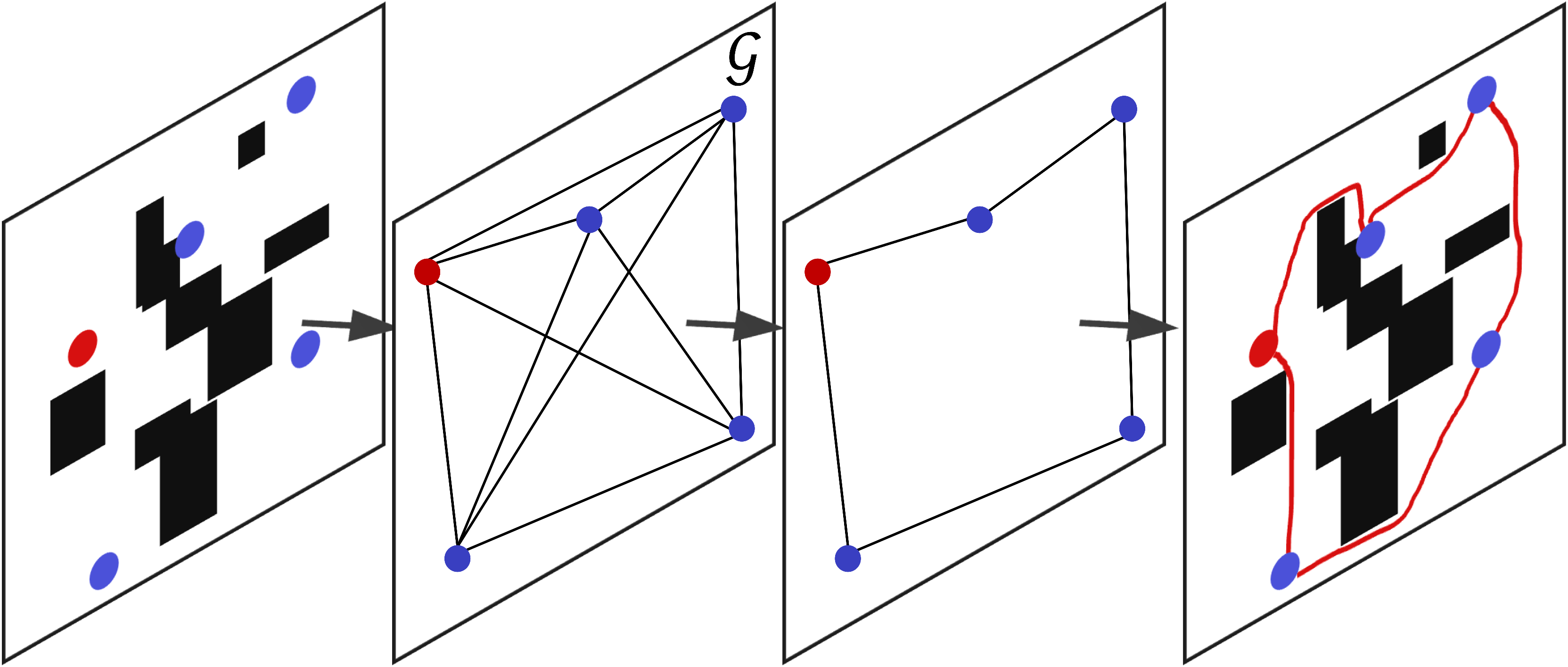}
	
	\caption{Illustration of the physical TSP to address the MGPF problem. The environment consists of black obstacles and feasible white space. An origin is denoted by a red node, while other goals are blue. An undirect complete graph $\mathcal{G}$ is established with weights $\mathcal{W}$ on edges $\mathcal{E}$ between all pairwise vertices $\mathcal{V}$. TSP planners perform to compute the visiting order for path planners to obtain a final path.}
	\label{fig1}
\end{figure}

As an extension problem of path planning, multi-goal path finding (MGPF) focuses on finding a feasible path to visit a sequence of goals in an obstacle-filled environment. Recently, this problem gains increasing attention in various real-world scenarios, such as data collection \cite{ref1,ref2,ref3}, pickup-and-delivery \cite{ref4},\cite{ref5}, and active perception \cite{ref6}. Unlike the path planning problem, MGPF can be decoupled as two sub-problems: finding the visiting order, and finding a path following the visiting order. Generally, the order is computed referring to the graph to minimize the visiting cost (i.e., path length). In this case, the MGPF can be generalized as a physical Travelling salesman problem (TSP) \cite{ref7} with an undirected complete connected graph. As illustrated in Fig. \ref{fig1}, multiple goals are denoted as vertices, while the edges represent the feasible paths between any pairwise goals. Weights on the edges are lengths of the feasible local paths between each pairwise vertices. MGPF algorithms use TSP planners \cite{ref8,ref9,ref10} to determine the visiting order for path planners, ultimately searching for a final path. As an NP hard problem \cite{ref11}, accurate approximations of the weights play a critical role in the maintenance of the optimality of the visiting order.

\begin{figure}[t]
	\centering
	\includegraphics[width=3.2 in]{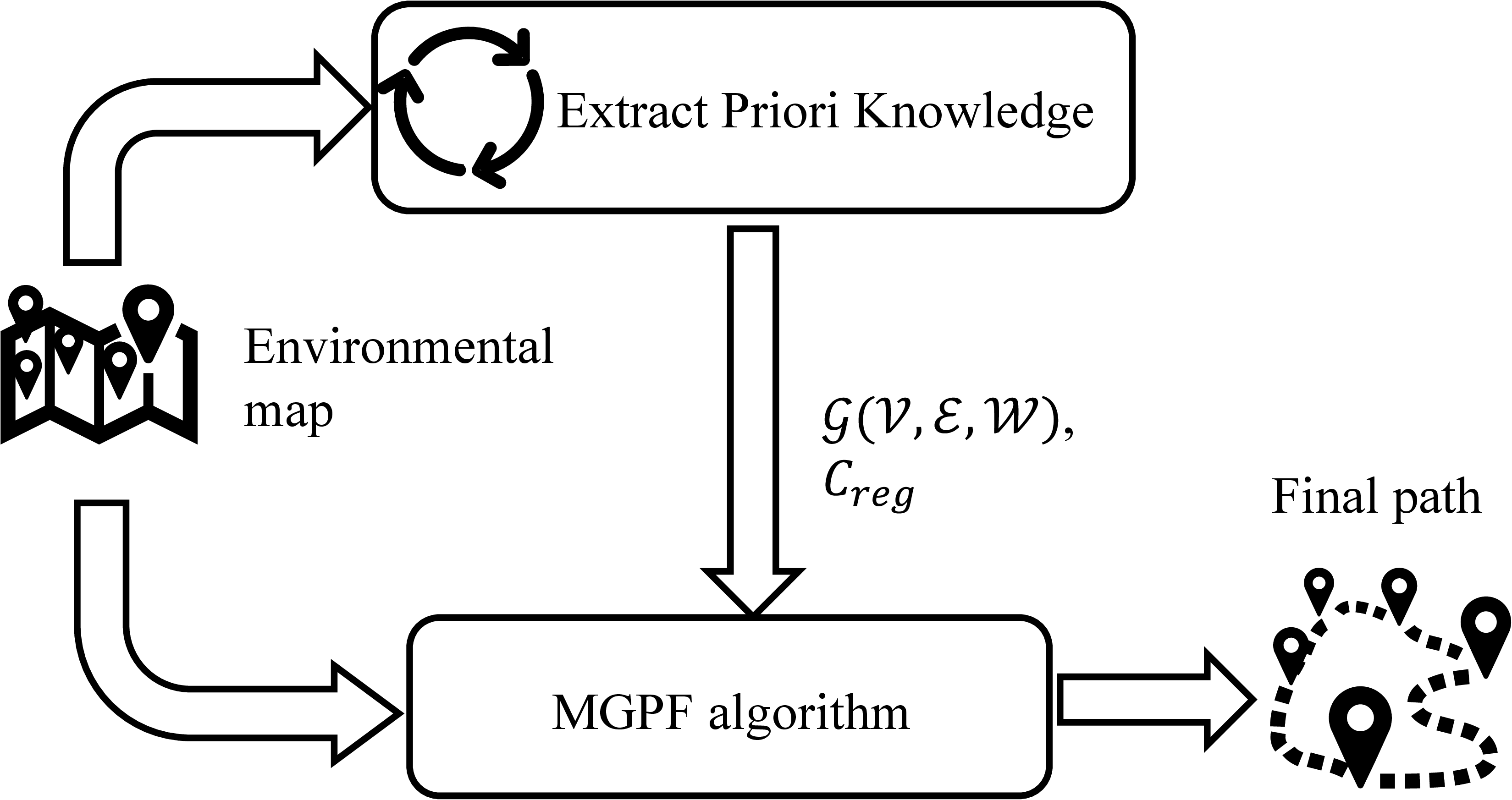}
	
	\caption{Illustration of the framework of the MGPF algorithms guided by prior knowledge. Prior knowledge are extrated to build a complete graph $\mathcal{G}(\mathcal{V},\mathcal{E},\mathcal{W})$ and a promising region $C_{reg}$.}
	\label{fig2}
\end{figure}

However, it is challenging to construct a complete connected graph with accurate weights since less prior knowledge of the feasible local paths is provided. A Euclidean distance can be calculated as a conservative estimate of the weight. Obviously, it is problematic to obtain the optimal visiting order. To obtain accurate weights, traditional approaches employ sampling-based path planners \cite{ref12,ref13,ref14} to explore the configuration space by generating samples rapidly without discretization. Best et al. \cite{ref15} construct a self-organizing map cooperating with the Rapidly-exploring Random Graph \cite{ref12} to find the favorable visiting order. Devaurs et al. \cite{ref16} introduce a geometric path planner based on a transition-based Rapidly-exploring Random Tree (RRT) \cite{ref17}, while low-level paths are computed to produce a high-quality path without a symbolic planner. Von{\'a}sek et al. \cite{ref18} utilize multiple trees to explore the space from all goals simultaneously until filling the whole space. Jano{\v{s}} et al. \cite{ref19} further enhance the path quality by rewiring and pruning with a priority queue for the priority growth of trees. Nevertheless, the approaches establish an incomplete weighted graph, thereby compromising the attainment of the optimal visiting order. Some approaches attempt to compute all weights between each pairwise vertices by sampling-based planners. In \cite{ref20}, Informed Steiner Trees utilizes samples to form a roadmap and de-emphasize the configuration regions with non-promising samples. A Steiner tree performs for a Mini-mum spanning tree in the complete  graph with lower bounds and upper bounds for edges. S*\cite{ref21} fuses the spirits of the Informed Steiner Trees with heuristic searching to improve efficiency. In \cite{ref18}, probabilistically complete methods based on RRT and probabilistic roadmap (PRM) \cite{ref14} are tested, while the planners run sequentially to connect all pairwise goals. Nevertheless, without prior knowledge of the local paths, it is time-consuming to precisely build the connections for all-pair vertices. So, how to efficiently obtain the desired weights with acceptable accuracy? What's more, sampling-based planners suffer from inherently uniform samples, which obstruct their further progress in MGPF. The uniform sampler generates samples randomly to explore the whole space, leading to unnecessary exploration. Besides, the uniform samples exacerbate the challenges for sampling-based planners in handling path planning with narrow passages.
In recent research \cite{ref22,ref23,ref24}, neural networks are used to extract the prior knowledge of the path between two vertices, which is represented by a promising region. Specifically, the region surely contains the optimal path, and subsequently performs as a heuristic to enhance the exploration efficiency by productive samples, especially in a complex environment. Inspired by this, we focus on using neural networks to extract the prior knowledge of the environments and path for estimating the weights, and enhance the quality of samples in \cite{ref25}. Fig. \ref{fig2} illustrates the framework of the approaches based on the prior knowledge. The extraction network acquires the environmental map to iteratively realize a prior construction of the complete weighted graph. The estimation of weights is regarded as a regression task to obtain an approximate length of the local path between pairwise vertices. Simultaneously, a segmented region is derived from an encoder-decoder architecture, which enables the path planner efficiently to explore the configuration space.

In this work, we continue this line of thinking, and design a novel multi-task learning model, namely Prior Knowledge Extraction (PKE), which consists of a segmentation task, a prediction task, and a regression task. For the segmentation task, a promising region that surely contains the optimal path between two vertices is produced, while the prediction task aims to produce a guideline in the region. Weights for the complete graph is estimated in the regression task the regression task. This model performs iteratively to obtain prior knowledge before incorporating an MGPF algorithm. Furthermore, we integrate the outputs of the PKE model with RRT as a novel approach, abbreviated to PKE-RRT, to solve the MGPF problem. Specifically, a TSP solver computes the visiting order first based on the complete weighted graph. The segmented region and guideline serve as heuristics to effectively guide the searching process before finding a closed and feasible path based on the visiting order. Finally, numerical experiment results demonstrate the efficiency and reliability of the proposed PKE-RRT approach in solving the MGPF problem, in respect of calculation time, solution cost, sample number, and success rate.

The rest of the paper is organized as follows. In Section \ref{sec2}, we give a definition of the MGPF problem, and a brief description of a sampling-based planner RRT is introduced. In Section \ref{sec3}, the details of the multi-task learning model PKE and MGPF algorithm PKE-RRT are presented, respectively. Numerical experiments are implemented based on seen and unseen scenarios in Section \ref{sec4}. Finally, we reach a conclusion, and highlight the main contribution of this research in Section \ref{sec5}.

\section{PRELIMINARIES}
\label{sec2}
In this section, we formulate the MGPF problem and offer a brief introduction to a sampling-based planner RRT.
\begin{figure*}[t]
	\centering
	\includegraphics[width=7 in]{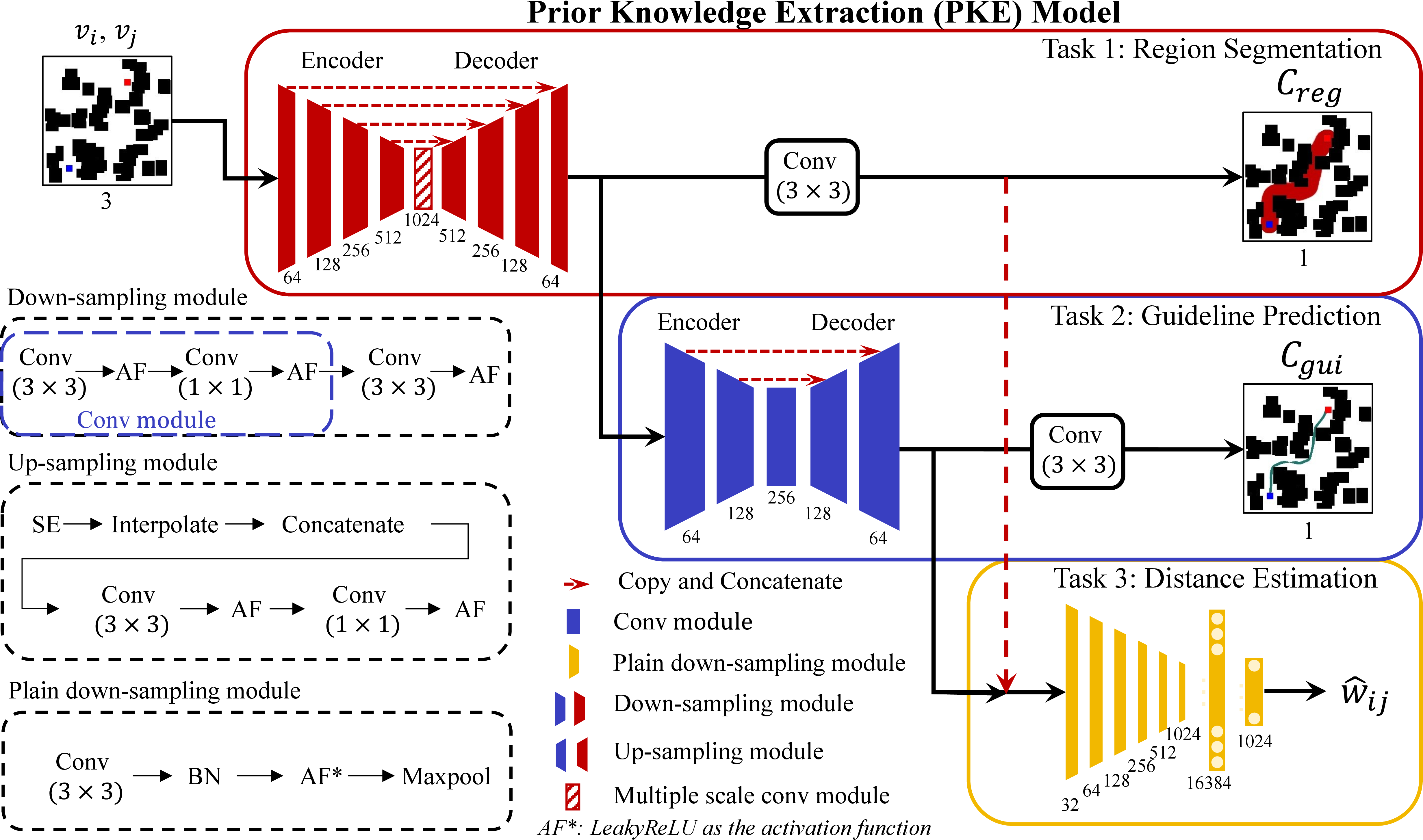}
	
	\caption{Architecture of the Prior Knowledge Extraction (PKE) model. The segmented region is denoted by $C_{reg}$ in red, while the predicted guideline is represented by $C_{gui}$ in green on the original map for better illustration. The channel number of the feature maps are represented by numerical.}
	\label{fig3}
\end{figure*}
\subsection{Problem Definition}
\label{sec2a}
As mentioned in Section \ref{sec1}, the MGPF problem can be generalized as a physical TSP. A closed and feasible path $\sigma=\cup_{i=1,...,N}\sigma_i$ is required to visit $N$ goals and return to an origin following a sequence $\mit\Psi=(\mit\psi_{i}):i\in[1,...,N+1]$ with the minimum cost. Different from the traditional TSP, each local path $\sigma_i$ is generated under the constraint of obstacle space $C_{obs}$. Let $C$ represent the configuration space, and $C_{free}$ denotes the free space. As shown in Fig. \ref{fig1}, the undirect complete graph $\mathcal{G}(\mathcal{V},\mathcal{E},\mathcal{W})$ is established according to the number of goals. Multiple goals are donated as vertices $\mathcal{V}=\left(v_i\right)\in C_{free}:i\in[1,\ldots,N]$, while a connection between pairwise vertices is denoted by edge $\mathcal{E}=\left(e_{ij}\right):i,j\in[1,\ldots,N]$, and $i\neq j$. A weight $w_{ij}\in\mathcal{W}$ is described as the length of a feasible local path to connect the corresponding vertices $v_i$ and $v_j$.  In this article, the solution cost is defined as the sum of the length of the local paths between visited vertices. The local length $\sigma_i$ is formulated as $c(\sigma_i)=\int_{0}^{1}{{||\sigma}_i(s)||ds}$, while $\sigma_i(s)$ represents feasible states on the local path, and $\sigma_i\left(0\right),\sigma_i(1)\in\mit{\Psi}$. The MGPF problem can be formulated as\\
\begin{equation}\label{eq1}
\sigma=arg\min({{\sum_{i=1}^{N}{c(\sigma_i)}}})
\end{equation}
\begin{equation}\label{eq2}\vspace{-1ex}
s.t.\psi_0=\psi_{N+1}
\end{equation}
\begin{equation}\label{eq3}\vspace{-1ex}
\forall\sigma_i\left(0\right),\sum_{j=1}^{N}f\left(\sigma_i\left(0\right),v_j\right)=1
\end{equation}
\begin{equation}\label{eq4}\vspace{-1ex}
\forall\sigma_i\left(1\right),\sum_{j=1}^{N}f\left(\sigma_i\left(1\right),v_j\right)=1
\end{equation}

\begin{equation}\label{eq5}\vspace{0ex}
	 \forall v_i,v_j\in\mathcal{V},f\left(v_i,v_j\right)=\left\{
	 \begin{array}{rcl}
1,&&if\; \sigma_i(0)=v_i,\sigma_i(1)=\\ &&\quad v_j,\text{and}\;\sigma_i\subset\sigma\\
0,&&\text{otherwise}
\end{array} \right.
\end{equation}
\begin{equation}\label{eq6}
u_i-u_j+N\ast f\left(v_i,v_j\right)\le N-1,\ 1\le i\neq j\le N
\end{equation}
(\ref{eq1}) guarantees the final solution is lowest-cost under the constraints. (\ref{eq2}) indicates the path starts and ends at the same vertex. (\ref{eq3}) and (\ref{eq4}) reveal that there are only one incoming local path and one outgoing local path for all vertices. (\ref{eq6}) indicates that there is only one closed path visiting all vertices, while $u_i$ is a dummy variable. It should be noted that the optimal solution is closely related to the visiting order $\mit{\Psi}$, which is derived from graph $\mathcal{G}(\mathcal{V},\mathcal{E},\mathcal{W})$. Thus, there is a reasonable and significant needs for accurate estimations of the weights $\mathcal{W}$, while $w_{ij}\approx\int_{0}^{1}{{||\sigma}_i(s)||ds}$, and $\sigma_i\left(0\right)=v_i,\sigma_i\left(1\right)=v_j$.

\subsection{Sampling-Based Planner RRT}
\label{sec2b}
Classic sampling-based planners employ samples to find the optimal solution, and the samples construct a geometric structure referring to certain growth rules. As a widely used algorithm, RRT employs a randomly exploring tree to search the feasible path incrementally. The tree structure is simply implemented, and its uniform expansion feature enables RRT available in various environments. As shown in lines 12-13 of Algorithm \ref{alg:alg1}, a new state is generated towards the random sample, and the feasibility and connectivity test perform to identify the state is available to be appended to the tree structure. The algorithm will be terminated if a feasible path is constructed from an origin to a goal. It is fundamental to promote the efficiency of the exploration with high-quality samples. As there are some narrow passages in $C_{free}$, it is computationally expensive to find a feasible path passing through the narrow passages. Previous studies \cite{ref22,ref23,ref24} have shown an effective perspective to aggregate neural networks with sampling-based planners. The neural networks extract the prior knowledge of the potential path to predict the distribution of effective samples. Thereby, the configuration space is shrunk into a promising region $C_{reg}$, allowing the exploration tree to quickly find a solution and avoid excessive sample attempts.

\section{PKE Model and PKE-RRT Algorithm}
\label{sec3}
In this section, we first present the proposed multi-task learning model PKE to extract the prior knowledge in Section \ref{sec3a}. Then, we introduce a two-stage training strategy to learn the joint features of multiple tasks in Section \ref{sec3b}. Finally, we describe the framework of the PKE-RRT for the MGPF problem in Section \ref{sec3c}.

\subsection{PKE Model}
\label{sec3a}
The PKE model is a multi-task learning model which is composed of three sub-models corresponding to three tasks.  As shown in Fig. \ref{fig3}, the input map is represented by a 3D RGB image, while $C_{obs}$ is black, and $C_{free}$ is white. Pairwise vertices are denoted by a red and a blue node, respectively. The outputs consist of a segmented region $C_{reg}$, a predicted guideline $C_{gui}$, and an estimated weight $w_{ij}$ of the local path between vertices $v_i$ and $v_j$. The outputs $C_{reg}$ and $C_{gui}$ are represented by binary maps with the same size as the input, while element 1 indicates a high probability of lying in a promising region or guideline. The regression result is returned as a number. The model is iteratively executed with different pair vertices to accomplish the complete weighted graph $\mathcal{G(V,E,W)}$.
\begin{figure}[t]
	\centering
	\includegraphics[width=3.2 in]{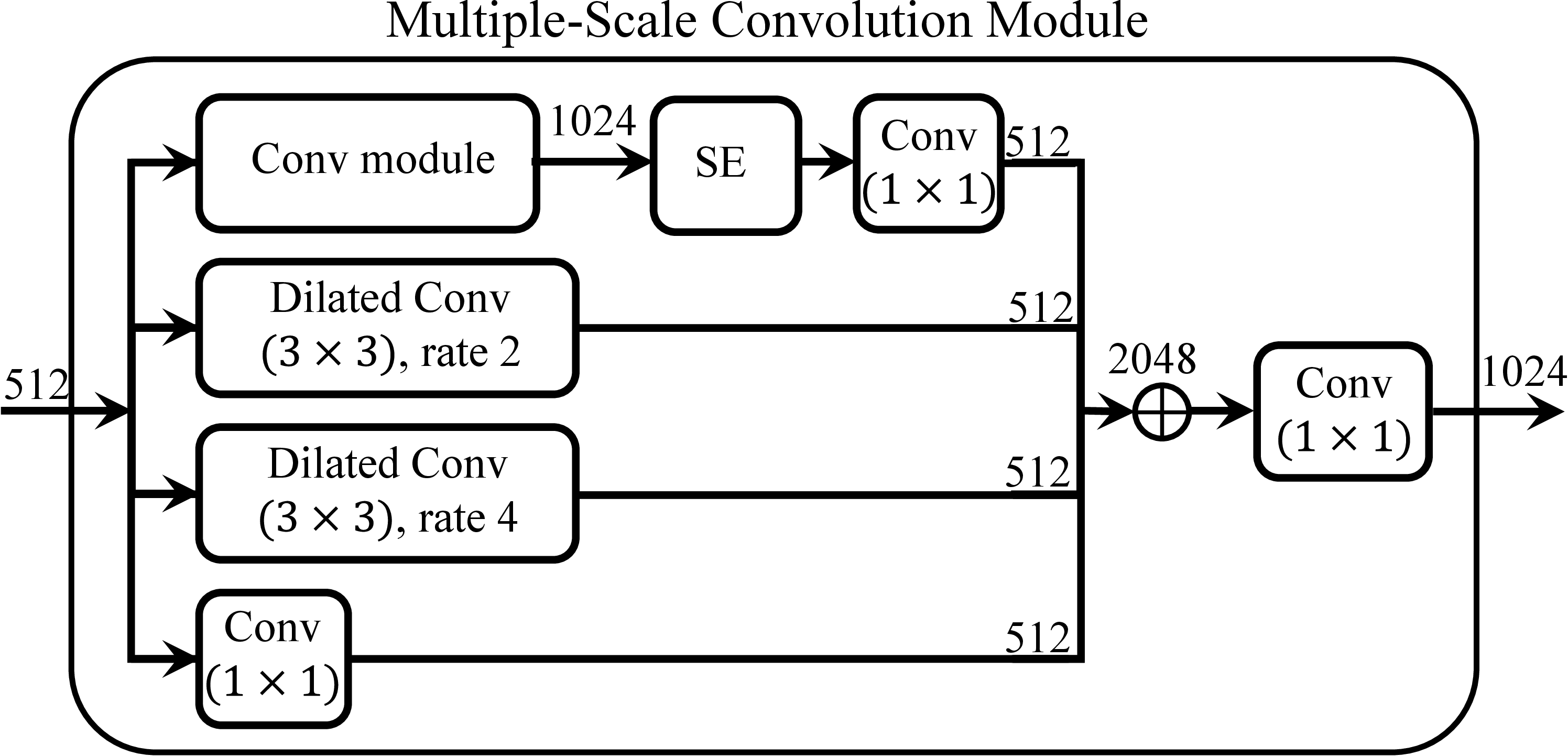}
	\caption{Structure of the multiple-scale convolution module. The channel number of the feature maps is represented by numerical.}
	\label{fig4}
\end{figure}
\begin{figure*}[b]
	\centering
	\subfloat[Separated training step.]{\includegraphics[width=3.1in]{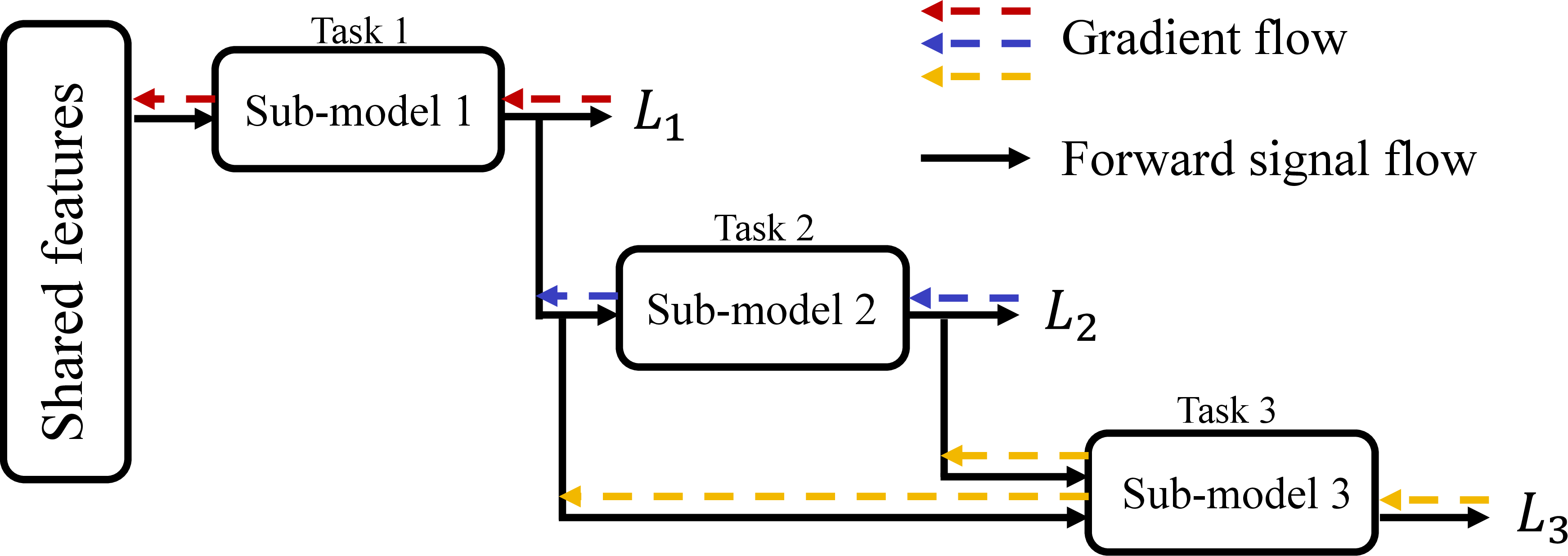}\label{fig5a}} 
	\hspace{0.1 in}
	\subfloat[Tuning training step.]{\includegraphics[width=3.1in]{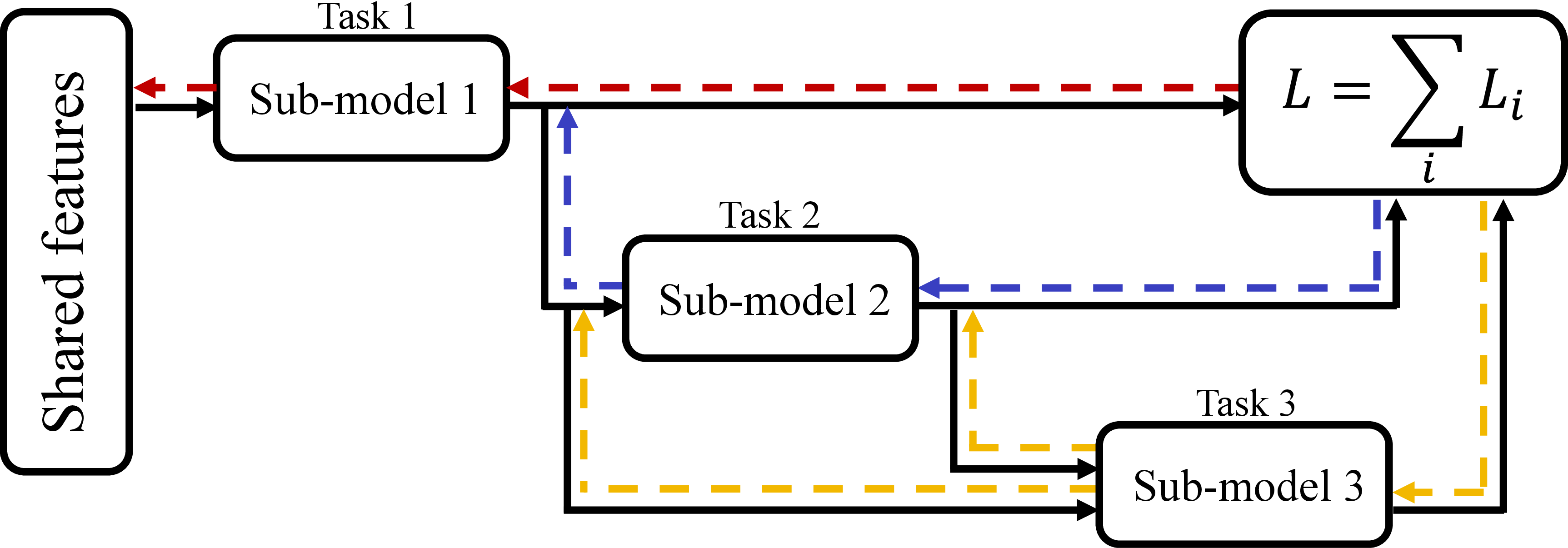}\label{fig5b}}
	\caption{Illustration of the two-step training strategy for joint learning.} 
	\label{fig5}
\end{figure*} 

 As shown in Fig. \ref{fig3}, the PKE model can be decomposed into three sub-models. The first sub-model adopts a classic U-Net-like structure with down-sampling modules, up-sampling modules, and a multiple-scale convolution module. The down-sampling module is to extract feature information, and subsequently decreases the resolution of the feature maps within the encoder. The reduction is implemented by a convolution operation with stride 2 to maintain more spatial relationships. Symmetrically, the up-sampling module recovers the spatial resolution of feature maps through interpolation, and incorporates the local feature maps with the corresponding feature maps from the encoder. To alleviate the semantic gaps between the encoder and decoder, a Squeeze-and-Excitation (SE) \cite{ref26} block is embedded before the concatenation. Additionally, a multiple-scale convolution module is designed to enlarge the receptive field, allowing the model to capture rich contextual information in the deep layer. The structure is shown in Fig. \ref{fig4}, while the feature maps are transferred into four distinct convolution operations. By progressively up-sampling and concatenating multi-level features, the sub-model finally generates a binary map denoted by $C_{reg}$. The second sub-model employs the same structure as the first sub-model. $C_{reg}$ is fed into the encoder, while the down-sampling module extracts the low-level and high-level information. The multiple-scale conv module is replaced by a convolution (denoted as Conv) module to reduce the calculation. The decoder predicts the possibility that each pixel belongs to the guideline at a pixel level. The output of the second sub-model is merged with $C_{reg}$ as inputs of the third sub-model. A multilayer perceptron sub-model is constructed with pre-processing by a plain down-sampling module, and then the final numeric output is produced as the estimated weight $\hat{\textit{w}}_{ij}$.

\subsection{Data Generation and Two-Step Training Strategy}
\label{sec3b}
We use the Multi-Goal Path Finding Dataset (MGPFD) in \cite{ref25} to train the model. The input data are RGB images with an origin and a goal. Rectangles of different sizes are randomly generated as obstacles. Labels of the regions are dilated from optimal solutions of RRT* \cite{ref12}. Moreover, the optimal path is annotated as the ground truth for the guideline prediction task. The annotated regions and guidelines are stored as binary images. The length of the optimal path is stored as the target weight, $w$. 

In the training process, individual loss functions are deployed for each learning task. For the segmentation task, we combine the focal loss \cite{ref27} and the Dice loss \cite{ref28}, which can be described as 
\begin{equation}\label{eq7}
\begin{split}
L_1=-{\sum\alpha}_t\left(1-\widehat{x}_t\right)^\gamma\log{\left(\widehat{x}_t\right)}+1-\frac{2\left|x\bigodot\hat{x}\right|}{\left|x\bigodot x\right|+\left|\hat{x}\bigodot\hat{x}\right|},
\end{split}
\end{equation}
while $	 \hat{x}_t=\{$
$\begin{array}{*{5}{rlr}}
	\hat{x}_{ij},&if\; x_{ij}=1\\
1-	\hat{x}_{ij},&otherwise
\end{array}$, and $\hat{x}_t$ is an output element at pixel position $(i,j)$ in $C_{reg}$. Besides, $\alpha_t$ is a coefficient to balance positive and negative data. In the latter term of (\ref{eq7}), $\bigodot$ represents a Hadamard product operation on the segmented output and the ground truth. For the prediction task, a loss $L_2$ replaces the focal loss with a binary cross entropy loss, which can be expressed as 
\begin{equation}\label{eq8}
	\begin{split}
		L_2=-{\sum}\log{\left(\widehat{x}_t\right)}+1-\frac{2\left|x\bigodot\hat{x}\right|}{\left|x\bigodot x\right|+\left|\hat{x}\bigodot\hat{x}\right|}.
	\end{split}
\end{equation}
For the regression task, a mean square error (MSE) is employed as $L_3$ to evaluate the disparity between the estimated weight and the target weight.
\begin{equation}\label{eq9}
L_3=\sum{(w-\hat{w})}^2
\end{equation}
Due to the difference in gradient and scale among the individual losses of multiple tasks \cite{ref29}, we conduct a two-step training strategy to adequately learn the features of various tasks. As illustrated in Fig. \ref{fig5}, the sub-models are trained separately in the first step. The loss $L_i:i=1,2,3$ is optimized individually and orderly with the gradient flows of the other sub-models frozen. Therefore, the multiple tasks can be trained well, involving the specific loss without the disturbance from the other losses. In the second step, the PKE model is trained unitedly to tune the parameters with a total loss, which can be expressed as $L_{total}=L_1+L_2+L_3$.
\begin{figure}[!b]
	
	\renewcommand{\algorithmicrequire}{\textbf{Input:}}
	\renewcommand{\algorithmicensure}{\textbf{Output:}}
	\removelatexerror
	\begin{algorithm}[H]
		\label{alg:alg1}
		\caption{PKE-RRT}
		
		\LinesNumbered 
		\KwIn{Map, Graph $\mathcal{G}$, Region: $C_{reg}$, Guideline: $C_{gui}$}
		\KwOut{$\sigma=\mathop{\cup}\limits_{i \in [1...N]}\sigma_i$}
		${\mit\Psi}=(\psi_i):i\in[1,...N+1] \leftarrow {\tt{Elkai}}{\texttt{(}}\mathcal{G}{\texttt{)}}$\; 
		$\sigma \leftarrow \emptyset$\;
		\For{$i=1,...,N$}{
			$\text{s}_\text{start},\text{s}_\text{goal} \leftarrow \psi_i, \psi_{i+1} $\;
			
			$V \gets \text{s}_\text{start};$  $E\gets \emptyset;$ $\textit{$T_i$} = (V,E)$\;
			\uIf{\tt{rand()}$>k_1$}{
				$\text{s}_\text{rand}\leftarrow$ {\tt{Nonuniform}($C_{gui}$)}\;}
			\ElseIf{\textup{\tt{rand()}}$<k_2$}{	$\text{s}_\text{rand}\leftarrow$ {\tt{Nonuniform}($C_{reg}$)}\;}\Else{$\text{s}_\text{rand}\leftarrow${\tt{UniformSample()}}\;}
			$	\text{s}_\text{nearest}	\gets {\tt{Nearest}}{{\texttt{(}}T_i,\text{s}_\text{rand}{\texttt{)}};}$\\
			$	\text{s}_\text{new}	\gets {\tt{Steer}}{{\texttt{(}}\text{s}_\textup{nearest},\text{s}_\text{rand}{\texttt{)}}}$\;
			\If{${\tt{Collisionfree}}$${\textup{\texttt{(}}}{\textup{s}}_{\textup{nearest}},\textup{s}_\textup{nearest}{\textup{\texttt{)}}}$}		{$V\leftarrow V\cup{\text{s}}_\text{new}$\;
				$E\leftarrow E\cup\{({\textup{s}}_\textup{new},\text{s}_\text{nearest})\}$\;}
			
			\If{${\textup{s}}_{\textup{new}}\in \textup{S}_\textup{goal}$}{$\sigma \leftarrow \sigma\cup \texttt{Extract}{\texttt{(}}T_i{\texttt{)}}$\;}
		}
		$\textbf{Return}\ {\sigma}$
		
	\end{algorithm}
\end{figure}

\subsection{PKE-RRT for MGPF Problem}
\label{sec3c}
As the PKE model is repeatedly deployed for all pairwise vertices in advance to make a prior extraction and construct a complete weighted graph $\mathcal{G}$ involving three tasks, the outputs can be integrated into sampling-based planners to address the MGPF problem efficiently. As illustrated in Algorithm \ref{alg:alg1}, the proposed PKE-RRT uses the complete weighted graph $\mathcal{G}$ to compute the visiting sequence $\mit\Psi$ via a traditional TSP solver Elkai \cite{ref9}. Then, a feasible path $\sigma$ is initialized. A local planner progressively searches for local paths between pairwise vertices in accordance with the sequence  $\mit\Psi$. Significantly, a hybrid sampling strategy is devised to improve search efficiency. We employ the segmented region $C_{reg}$ and the guideline $C_{reg}$ as heuristics to realize a nonuniform sampler, which generates effective samples in $C_{reg}$ or $C_{gui}$. To guarantee the probabilistic completeness of the algorithm, a uniform sampler is deployed with a probability. As the sample number increases to infinity, the hybrid strategy enables the whole configuration space to be fully filled with samples. Consequently, the PKE-RRT maintains the probabilistic completeness of the RRT. The local tree \textit{$T_i$} grows until a feasible path is found to connect the pairwise vertices. Subsequently, a local path $\mit\sigma_i$ is extracted from the local tree and added to $\mit\sigma$. The PKE-RRT iteratively searches for the local paths until a closed path $\sigma$ is found.

\begin{figure}[t]
	\centering
	\includegraphics[width=3.4 in]{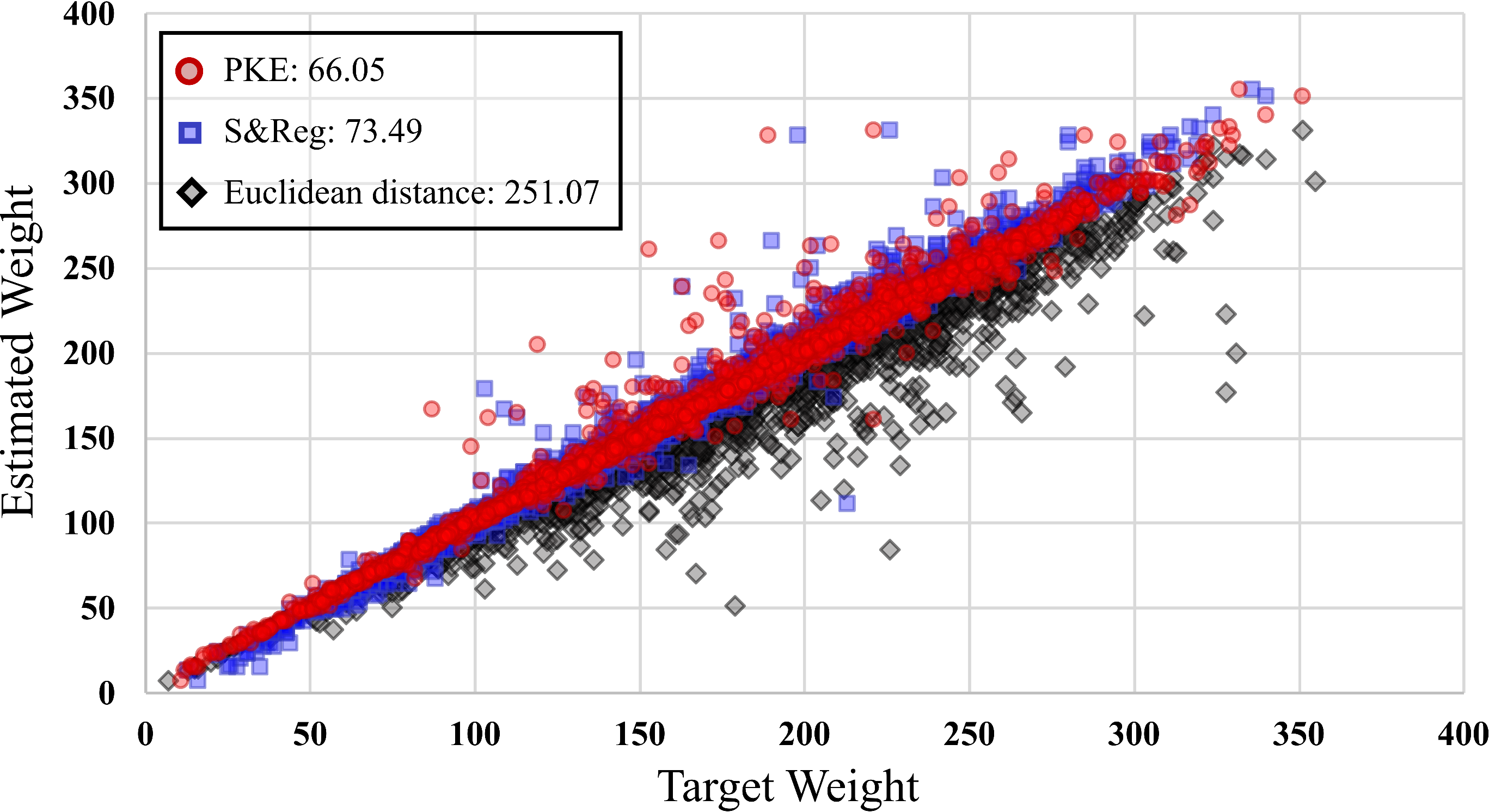}
	
	\caption{Regression performance of the PKE model on test dataset. The distribution shows the accuracy of the regression results for estimating the weight, which is assessed by the MSE.}
	\label{fig6}
\end{figure}
\begin{table}[t]
	\caption{Performance of the PKE Model on Test Dataset}
	\label{tab1}
	\begin{center}
		
		\begin{tabular}{ccccc}
			
			\hline
			\rule{0pt}{8pt}
			& \makecell[c]{IOU\\(Region)} & \makecell[c]{F1\\(Guideline)} & \makecell[c]{MSE\\(Regression)} & \makecell[c]{Frame\\(img/sec)}\\
			\hline
			\rule{0pt}{8pt}
			PKE & 0.8289 &0.6344&66.05&48.9\\
			\hline
			\rule{0pt}{8pt}
			S\&Reg &0.8160&-&73.49&54.2\\
			\hline

		\end{tabular}
	\end{center}
\end{table}
\begin{figure}[t]
	\centering
	\includegraphics[width=3.4 in]{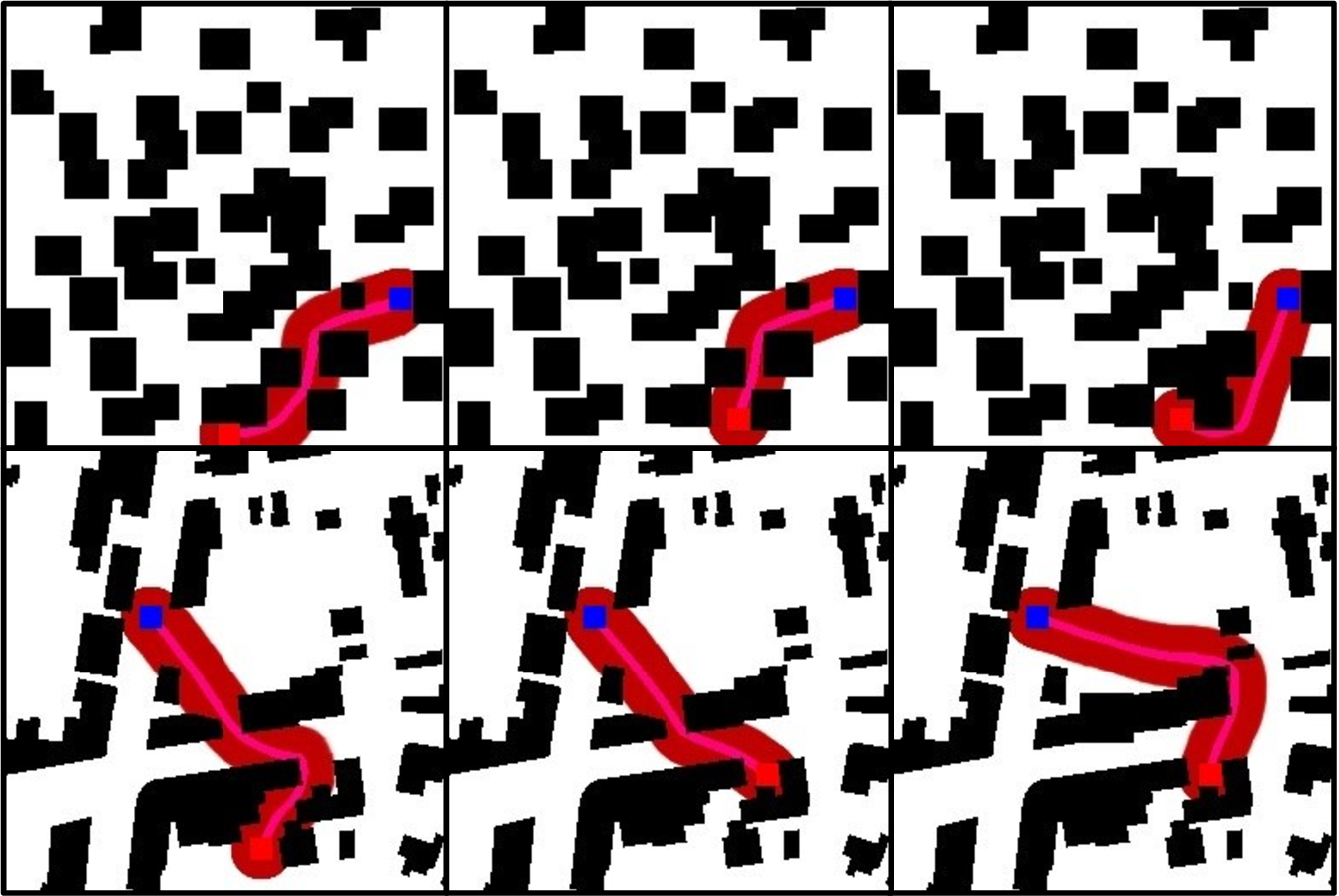}
	\caption{Online extraction results using the PKE model on static and changing scenarios. The origin and goal are denoted by a red node and a blue node, respectively. The dark red area is the promising region, and the bright red line is the guideline.}
	\label{fig7}
\end{figure}

\section{SIMULATION EXPERIMENTS}
\label{sec4}
In this section, we implement several numerical simulations to validate the performance of the PKE model and the PKE-RRT for MGPF. In Section \ref{sec4a}, we demonstrate the measurable improvement in the segmentation and regression results. In Section \ref{sec4b}, we conduct MGPF simulations based on the MGPFD, including two categories: seen and unseen scenarios. The seen scenarios stem from the test dataset, while the unseen scenarios are generated from the \cite{ref30} or filled with small and dense obstacles.

\subsection{Performance of the PKE Model}
\label{sec4a}

First, we implement experiments on the test dataset to verify the improvement of the PKE model in multi-task learning. The multiple tasks results are shown in TABLE \ref{tab1}, which is compared with a learning-based model S$\&$Reg. For the segmentation results, the PKE model reaches a 1.29$\%$ improvement in the Intersection over Union (IOU) score. The regression results are assessed by the MSE, and PKE model outperforms S$\&$Reg with an improvement of approximately 10.12$\%$. In Fig. \ref{fig6}, distributions of the PKE regression results are more concentrated on the line of equality compared with the S$\&$Reg and Euclidean distance. This reveals the PKE model can obtain more accurate estimated weights, which impact the optimality of the visiting order. Here, Euclidean distance is computed as the vertices are known in advance, which suggests the need for further refinement. It is worthy of note that the PKE model can achieve 48.9 images per second to meet the requirement of online extraction in a dynamic environment. In Fig. \ref{fig7}, we display some results from the seen and unseen scenarios. We assume the extraction of the PKE model is required for robots to lead the local path finding under three circumstances: (1) at the initial time while obstacles are static; (2) the robot is moving, and the origin is changing; (3) new obstacles appear and block the original path. The extraction results indicate that the PKE model can provide available knowledge as heuristics both in static and changing environments.
\begin{figure} [t]
	\footnotesize	
	\centering
	\setcounter {subfigure} {0}{
		\includegraphics[width=3.4    in]{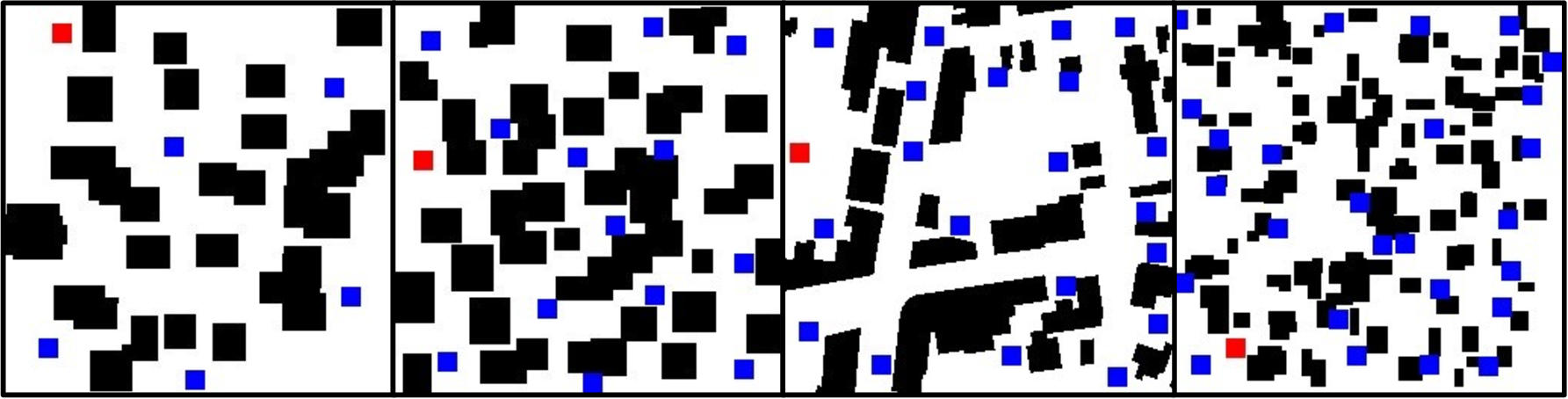}}\\
	\setcounter {subfigure} {1}{
		\includegraphics[width=3.4    in]{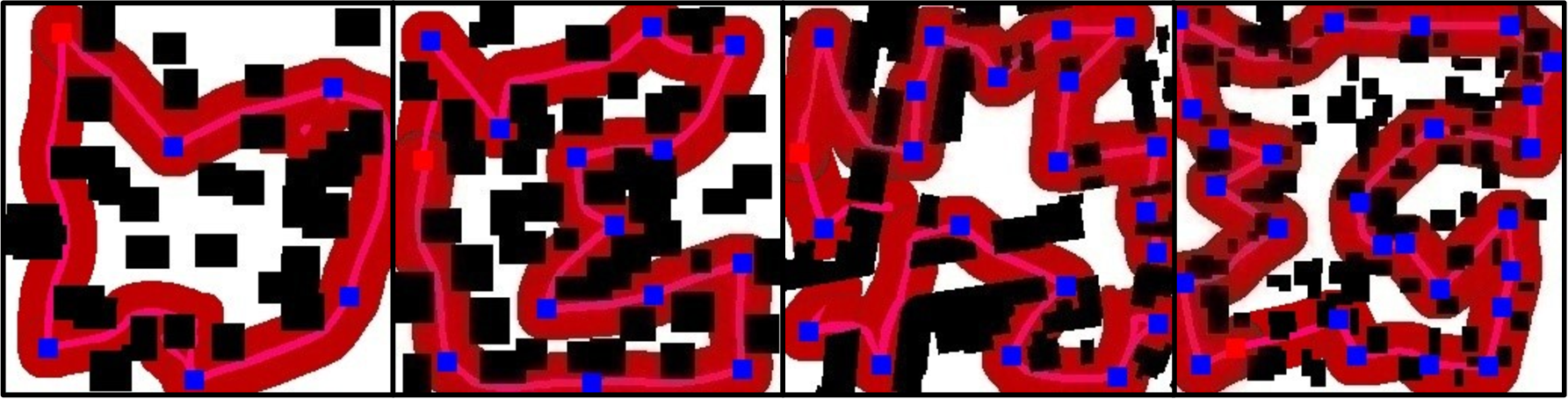}}\\
	\setcounter {subfigure} {2}{
		\includegraphics[width=3.4    in]{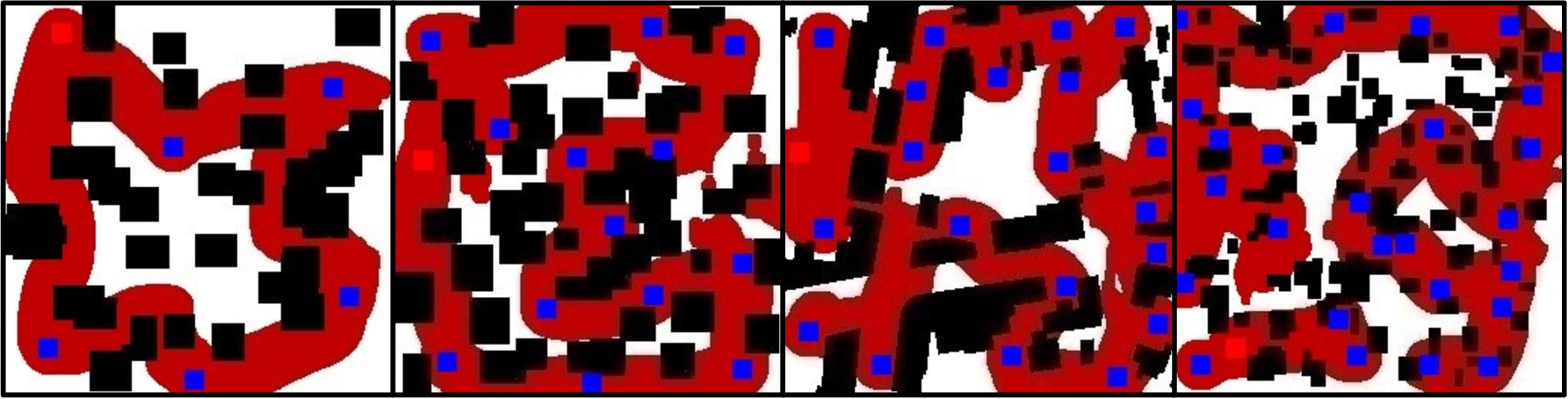}}\\
	\caption{Various scenarios for MGPF simulations. From left to right, there are seen scenario 1(S1), seen scenario 2 (S2), unseen scenario 1 (U1), and unseen scenario 2 (U2). The second and third rows are the prior knowledge extracted by the PKE and S$\&$Reg. Guidelines are denoted by bright red lines, and segmented regions are marked in dark red.}
	\label{fig8}
\end{figure}
\begin{figure}[t]
	\centering
		\includegraphics[width=3.1 in]{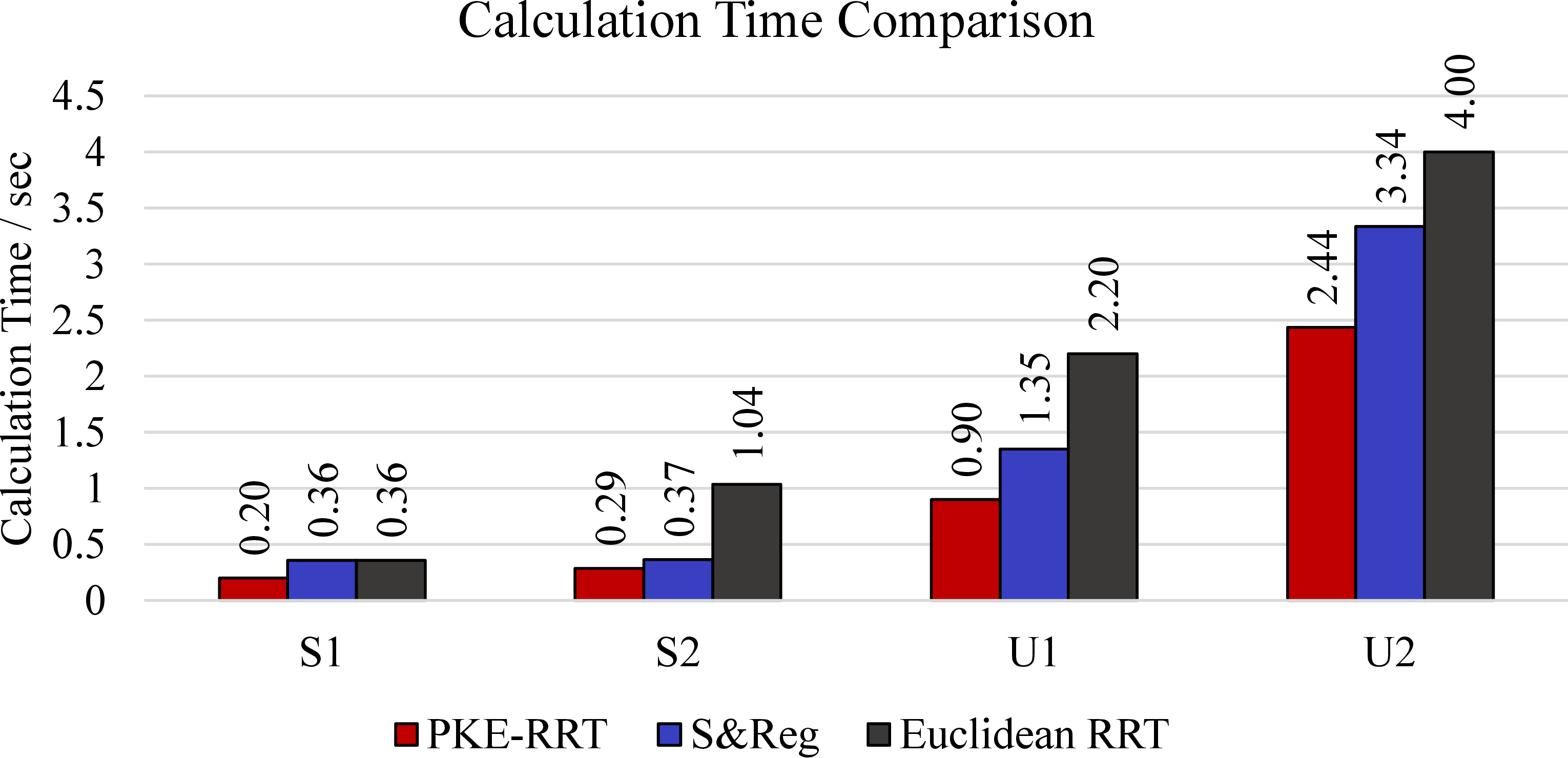}
	\caption{Comparison results of the calculation time in different scenarios.} 
	\label{fig9}
\end{figure} 
\begin{figure}[t]
	\centering
		\includegraphics[width=3.1 in]{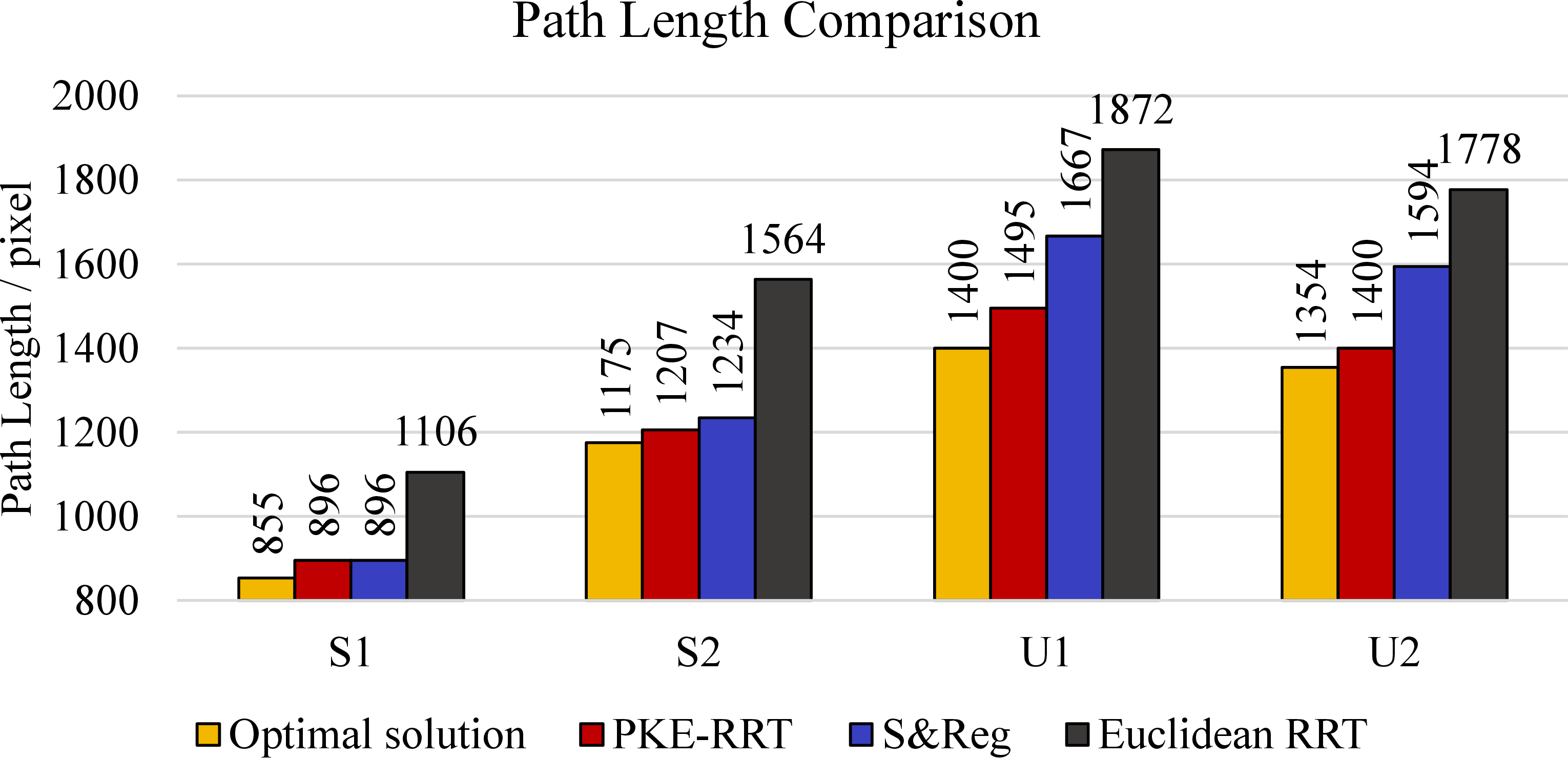}
	\caption{Comparison results of the path length in different scenarios.} 
		\label{fig10}
\end{figure} 

\subsection{Performance of the PKE-RRT}
\label{sec4b}
To demonstrate the efficiency of the PKE-RRT, we conduct experiments based on the scenarios from the seen and unseen categories in the MGPFD. As shown in Fig. \ref{fig8}, the scenarios are designated according to the categories with a 256x256 size. Extraction results of the PKE and the S$\&$Reg are shown in the second and third rows. It is noted that we only exhibit the results referring to the visiting sequence instead of all possible connections between vertices. The comparison results indicate that the guideline can successfully connect two vertices to lead the search in most cases, and the regions surely cover the optimal local path. Besides, the extraction of the PKE model is more accurate than the S$\&$Reg. The latter produces some small and dissociated regions to mislead the search in the follow-up process. Interestingly, the position and shape of the regions are quite different from the S$\&$Reg in scenario S1, although the visiting order is identical. This phenomenon is possibly attributed to close path lengths for going right or left, which results in a distinct quality of the final solutions of the scenario S1 to some extent.

\begin{figure*}[htbp]
	\centering
	\includegraphics[width=7 in]{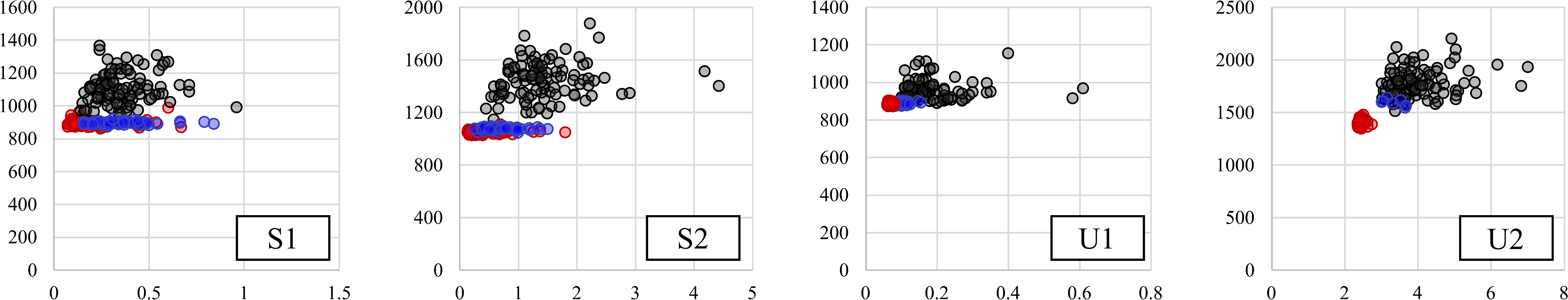}
	\caption{Distributions of the calculation time and path length based on the successful cases. PKE-RRT: Red. S\&Reg: Blue. Euclidean RRT: Black.} 
	\label{fig11}
\end{figure*}

We compare the PKE-RRT with S$\&$Reg, and Euclidean RRT, which utilizes the Euclidean distance as the weights of the graph. The Elkai TSP solver is employed to compute the visiting order. In the pathfinding process, the maximum sample number is set to 2000, and the algorithm is terminated if the algorithm generates samples more than 2000 or finds a feasible path. All algorithms are running 100 times. Besides, we utilize RRT* to compute the shortest path between all pairwise vertices for the complete connected graph, which is denoted as the optimal solution. Results of the calculation time and the path length are displayed in Fig. \ref{fig9} and \ref{fig10}. For all scenarios, PKE-RRT outperforms the S$\&$Reg and Euclidean RRT with less calculation time. Specifically, compared with the PKE-RRT, the calculation time is increased a lot for S$\&$Reg in unseen scenarios due to the disconnected regions. In Fig. \ref{fig10}, the PKE-RRT generates the closest paths to the optimal solution. For the Euclidean RRT, the non-optimal visiting order partially impacts the length of the final path. Moreover, we illustrate the distributions of the calculation time and the path length based on the successful searching cases in Fig. \ref{fig11}. We can see that the PKE-RRT results (red nodes) are concentrated in a small range of area, demonstrating the reliability of the PKE-RRT to solve the MGPF problem.

\begin{table*}[t]
	\caption{Sample Number and Success Rate Results}
	\label{tab2}
	\begin{center}
		
		\begin{tabular}{c|ccc|ccc}
			
			\hline
			\rule{0pt}{8pt}
			&& Sample number&&& Success rate &\\
			\cline{2-7}
			\rule{0pt}{8pt}
			& Euclidean RRT &S\&Reg&PKE-RRT& Euclidean RRT &S\&Reg&PKE-RRT\\
						\hline
			\rule{0pt}{8pt}
			S1 &1543&4498&600&100&86&100\\
			\hline
			\rule{0pt}{8pt}
			S2 &3779&2496&412&100&100&100\\
			\hline
			\rule{0pt}{8pt}
			U1 &3756&7119&609&100&70&100\\
			\hline
			\rule{0pt}{8pt}
			U2 &4394&12494&822&100&7&100\\
			\hline
		\end{tabular}
	\end{center}
\end{table*} 
\begin{figure} [t]
	\footnotesize	
	\setcounter {subfigure} {0} {
		\includegraphics[width=3.2    in]{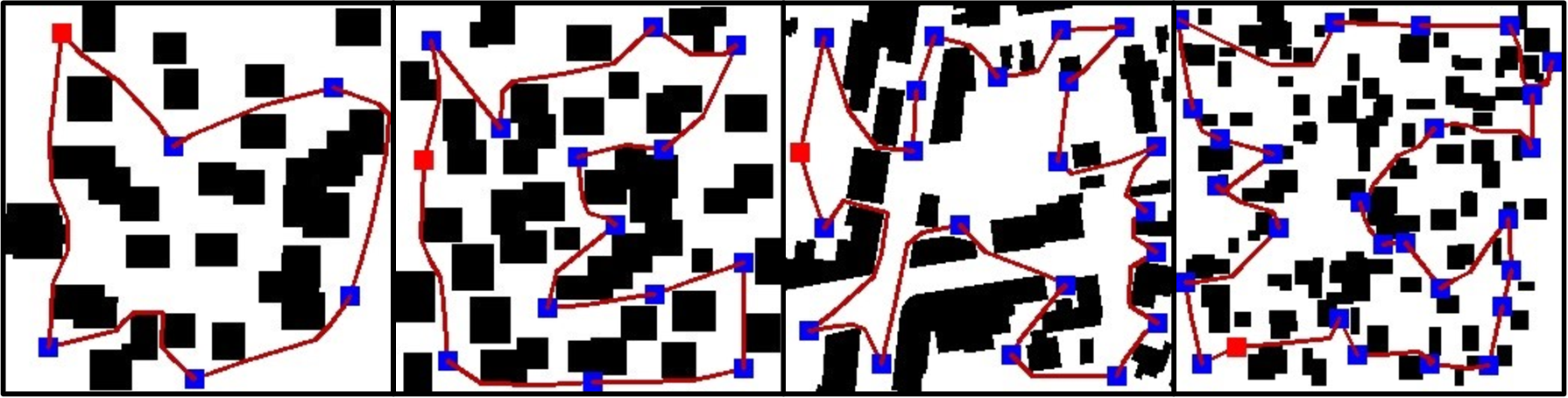}\label{fig10:a}}\\
	\setcounter {subfigure} {1} {
		\includegraphics[width=3.2    in]{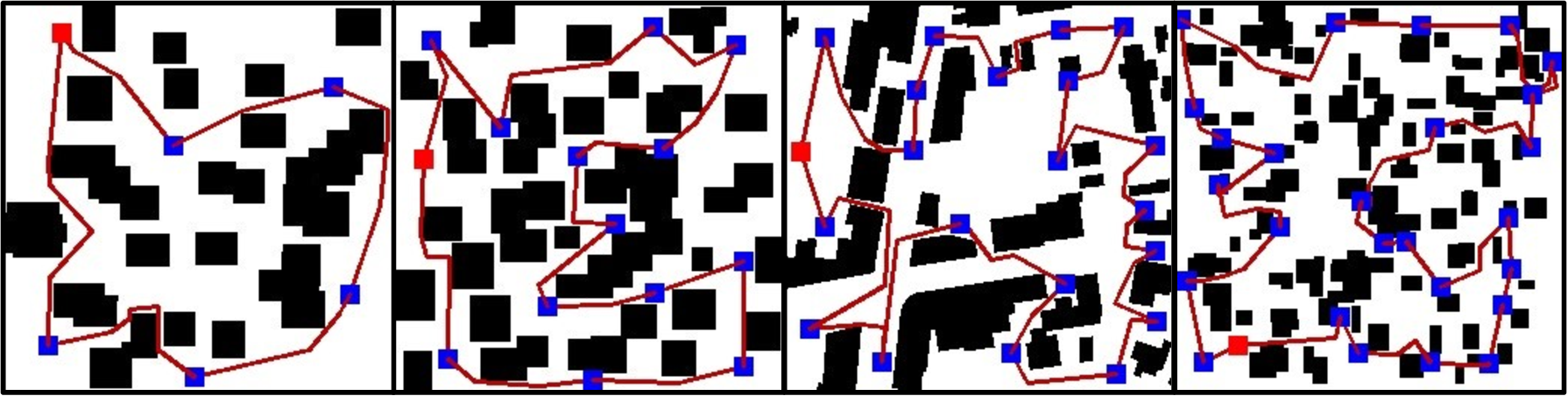} \label{fig10:b}}\\
	\setcounter {subfigure} {2} {
		\includegraphics[width=3.2    in]{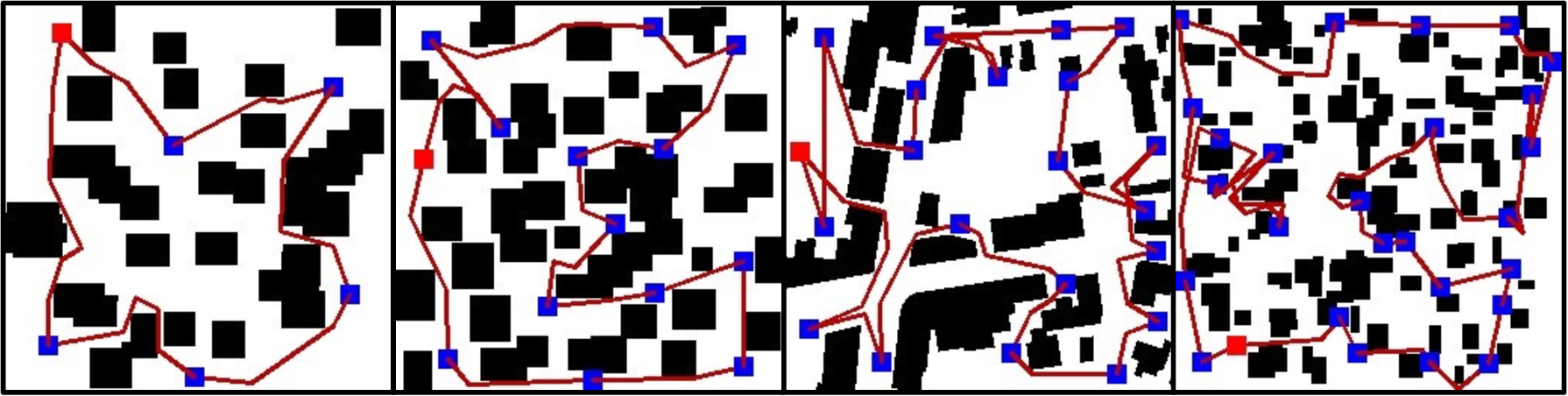} \label{fig10:c}}\\
	\setcounter {subfigure} {3} {
		\includegraphics[width=3.2    in]{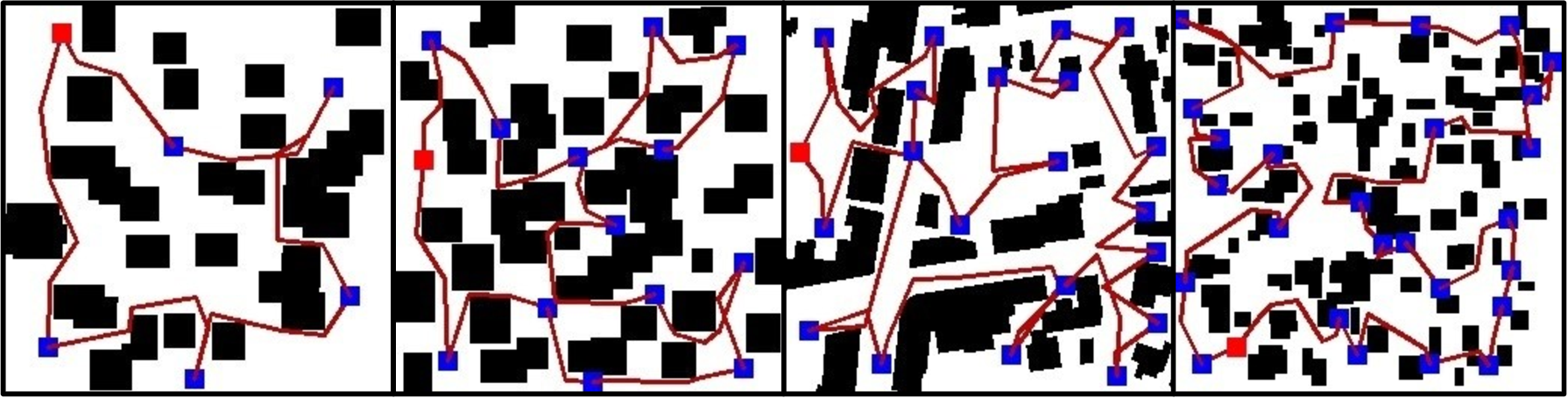} \label{fig10:c}}\\
	\caption{Multi-goal path planning results in S1, S2, U1, and U2. From top to bottom, there are the results of the optimal solution, PKE-RRT, S\&Reg, and Euclidean RRT.}
	\label{fig12} 
\end{figure}

In TABLE \ref{tab2}, we compared the sample number and the success rate. Here, a success case is defined as successfully finding a closed and feasible path according to the visiting order. PKE-RRT takes advantage of the accurate segmentation results and the guidance of the guideline, resulting in the fewest samples to find the feasible path. Besides, the success rate of the PKE-RRT verifies the reliability both in seen and unseen scenarios. The S$\&$Reg method exhibits a low success rate, particularly in the scenario U2, due to an inaccuracy in the identified regions and a lack of possibilistic completeness. The visualization results are exhibited in Fig. \ref{fig12}. It is worth noting that the visiting orders from all algorithms are the same as the optimal solution in scenarios S1. However, in the other scenarios, the results of the S$\&$Reg and the Euclidean RRT reveal a limited availability to find the optimal visiting order, while the PKE-RRT provides the optimal sequence with lowest cost. 

As sampling-based algorithms for the MGPF problem, SFF and SFF* explore the configuration space by multiple trees rooted at goals until the configuration space is fully filled with the trees. However, the results of the SFF and SFF* are not displayed due to their incomplete graph as well as the long calculation time. The multiple trees fail to connect each other, which cannot meet the requirement of (\ref{eq2}), (\ref{eq3}), (\ref{eq4}), and (\ref{eq6}). Furthermore, the elapsed time of over 35 seconds for solving the MPGF problem in scenario S1 indicates potential room for further improvement.

\section{CONCLUSIONS}
\label{sec5}
In this work, we aimed to extract prior knowledge for solving the MGPF (Multi-Goal Path Finding) problem efficiently. To achieve this, we designed the PKE (Prior Knowledge Extraction) model, a multi-task learning framework that facilitates the extraction of prior knowledge related to local paths between pairwise vertices. This extracted knowledge can be seamlessly integrated with sampling-based planners. The key components of the prior knowledge included a segmented region, a guideline, and an estimated length of the local path, providing an accurate weight for the generalized graph. Besides, a two-stage training strategy was devised to promote the joint learning. We then applied the PKE-RRT algorithm, which effectively utilizes the extracted prior knowledge, to address the MGPF problem. The visiting order was ensured to be optimal by computing the complete graph with accurate weights. Subsequently, the promising region and guideline served as heuristics, effectively reducing the configuration space, and accelerating the pathfinding process. Simulation experiments demonstrated the advantages of the PKE-RRT, including reduced calculation time, shorter path length, decreased sample number, and improved success rate. Furthermore, the performance of the PKE-RRT algorithm was evaluated on both seen and unseen scenarios, demonstrating its efficiency and reliability in solving the MGPF problem. 

Further research can involve investigating the underlying factors contributing to the phenomenon that distinct regions are generated when confronted with paths of similar lengths. Additionally, exploring extensions of the PKE model to other sampling-based planners is a promising research direction for further advancements in the field.





\section*{ACKNOWLEDGMENT}

This work was supported by JST SPRING, Grant Number JPMJSP2128.



\end{document}